\documentclass[letterpaper, 10 pt, conference]{ieeeconf}
\IEEEoverridecommandlockouts
\usepackage{amsmath,amsfonts}
\usepackage{algorithmic}
\usepackage{array}
\usepackage[caption=false,font=normalsize,labelfont=sf,textfont=sf]{subfig}
\usepackage{textcomp}
\usepackage{stfloats}
\usepackage{url}
\usepackage{verbatim}
\usepackage{graphicx}
\def\BibTeX{{\rm B\kern-.05em{\sc i\kern-.025em b}\kern-.08em
    T\kern-.1667em\lower.7ex\hbox{E}\kern-.125emX}}
\usepackage{balance}
\usepackage{subfloat}
\usepackage[graphicx]{realboxes}
\usepackage{booktabs}
\usepackage{algorithm}
\usepackage{hyperref}
\usepackage{adjustbox}
\usepackage{xcolor}
\let\labelindent\relax
\usepackage{enumitem}
\usepackage{cite}
\usepackage{listings}
\usepackage{xspace}

\hypersetup{
    colorlinks,
    pagebackref=true,
    linkcolor={red!50!red},
    citecolor={citecolor!50!citecolor},
    urlcolor={blue!80!black}
}

\definecolor{citecolor}{RGB}{30,130,255}
\newcommand{\code}[1]{\texttt{#1}}
\newcommand{\ourmethod}{{\code{Hamlet}}\xspace}

\begin{document}
\title{\LARGE \bf 
Integrating Learning-Based Manipulation and Physics-Based Locomotion for Whole-Body Badminton Robot Control}

\author{Haochen Wang$^{1}$, Zhiwei Shi$^{1}$, Chengxi Zhu$^{1}$, Yafei Qiao$^{1}$, Cheng Zhang$^{2}$, \\ Fan Yang$^{3}$, Pengjie Ren$^{1}$, Lan Lu$^{4}$$^{\dagger}$, and Dong Xuan$^{1}$
\thanks{\raggedright $^{1}$ School of Computer Science and Technology, Shandong University
}
\thanks{\raggedright $^{2}$ Robotics Institute, Carnegie Mellon University
}
\thanks{\raggedright $^{3}$ DeepCode Robotics
}
\thanks{\raggedright $^{4}$ Department of Sports, Shanghai Jiao Tong University
}
\thanks{\raggedright $^{\dagger}$ Corresponding author: \texttt{lulan871017@hotmail.com}
}
}

\maketitle

\thispagestyle{empty}
\pagestyle{empty}

\maketitle

\begin{abstract}
Learning-based methods, such as imitation learning (IL) and reinforcement learning (RL), can produce excel control policies over challenging agile robot tasks, such as sports robot.
However, no existing work has harmonized learning-based policy with model-based methods to reduce training complexity and ensure the safety and stability for agile badminton robot control.
In this paper, we introduce \ourmethod, a novel hybrid control system for agile badminton robots.
Specifically, we propose a model-based strategy for chassis locomotion which provides a base for arm policy. 
We introduce a physics-informed ``IL+RL'' training framework for learning-based arm policy. 
In this train framework, a model-based strategy with privileged information is used to guide arm policy training during both IL and RL phases. 
In addition, we train the critic model during IL phase to alleviate the performance drop issue when transitioning from IL to RL.
We present results on our self-engineered badminton robot, achieving 94.5{\%} success rate against the serving machine and 90.7{\%} success rate against human players.
Our system can be easily generalized to other agile mobile manipulation tasks such as agile catching and table tennis.
Our project website: \url{https://dreamstarring.github.io/HAMLET/}.
\end{abstract}

\section{Introduction}

Badminton is a competitive sport that requires high-speed reactions.
Participants are required to intercept high-velocity ball, reaching speeds up to 426 km/h, under an approximate one-second temporal constraint. 
The complexity of synchronizing visual and motor responses in such scenarios extends to robotics.
Here, the objective is the prompt unification of perception, decision-making, and actuation to accurately return the shuttlecock. 
Additionally, robotic systems must surmount the challenge of modeling the dynamics of flexible objects like shuttlecocks.

We develop a real robot, consisting of an omnidirectional chassis and an arm, to execute badminton strokes within a badminton court.
There are already some badminton robots.
{Kengoro}~\cite{kawaharazuka2017human}, a humanoid robot, demonstrates racket swinging but lacks genuine badminton play.
{Robomintoner}~\cite{web:robomintoner} achieves this yet with a simplistic arm mechanism limiting complex strokes like smash. 
In contrast, our intricate arm design facilitates a more diversified and agile operational range than {Robomintor}.
There are also some works \cite{abeyruwan2023agile, abeyruwan2023sim2real, ding2022learning, buchler2022learning, gao2020robotic,d2024achieving, ji2023opponent, zhang2018real, wang2023table}
similar to badminton robots.
However, given the requirement of swiftly navigating in a badminton court, implementing linear axes as used is challenging.

\begin{figure}[t]
\centering
\vspace{3mm}
\includegraphics[width=0.98\linewidth]{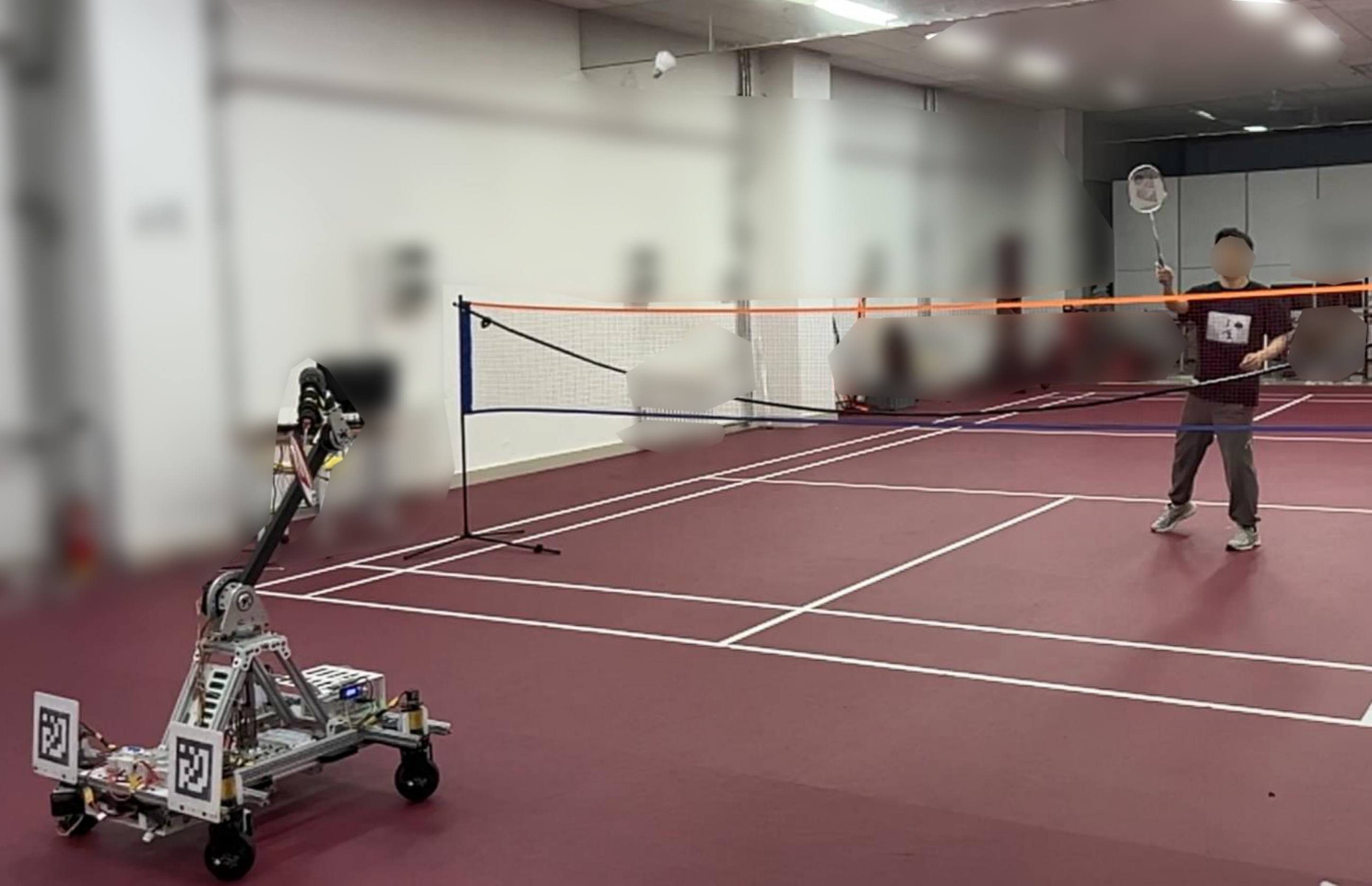}
\caption{\textbf{Agile badminton robot system.} The badminton robot on the left, which consists of an omnidirectional chassis and a 5-DOF arm, is playing badminton against a human on the right. The robot can move flexibly in the court and rapidly swing the racket to return the ball.}

\label{whole system figure in real world}
\end{figure}

Controlling such robots broadly involves two strategies: model-based and learning-based.
Model-based strategies \cite{mizuno2019development, yang2022varsm} offer both stability and safety.
However, they depend heavily on accurate environmental models. 
This reliance poses distinct challenges when regarding changes in the environment. 
Additionally, these strategies often require complex manual adjustments. 
This complexity further compromises their environmental robustness and flexibility.
On the other hand, 
learning-based strategies \cite{abeyruwan2023agile, abeyruwan2023sim2real, ding2022learning, buchler2022learning, gao2020robotic,d2024achieving, DBLP:conf/corl/FuCP22} show a higher success rate for tasks, particularly within the training data distribution. 
They also adapt more effectively to the environment. 
Nonetheless, they face particular difficulties, especially in policy network convergence. 
Another significant issue is the unexplainability of neural networks, which poses potential safety risks.
Merging model-based and learning-based strategies to leverage their respective strengths presents an avenue worth exploring.
\cite{abeyruwan2023agile, dong2020catch, pandala2022robust, ma2022combining} compare or integrate these two approaches.
However, they primarily employ learning-based policies to assist physical strategies.
And to the best of our knowledge, there is currently no such existing hybrid control system specifically designed for badminton robots.

Reinforcement learning is often utilized to train learning-based policies.
These policies can potentially surpass human-level performance. 
However, the time-consuming nature of RL training and challenges posed by tasks with sparse rewards are significant. 
Hence, an increasing number of works \cite{ramrakhya2023pirlnav, nair2020awac, lu2021aw, kalashnikov2018qt, rajeswaran2017learning, zhao2023learning, fu2024mobile, yu2023actor, wexler2022analyzing, DBLP:conf/rss/ZhangKDGS23} use offline data for the warming up of reinforcement learning methods or get control policy directly. 
Yet, the confined range of offline data distribution and potential performance drops during the transition from offline to online learning call for better solutions. 
For agile tasks, such as badminton robots, collecting human demonstrative offline data is challenging.
It makes us need to try necessitating alternative avenues for model warming up.

We propose \ourmethod, a novel hybrid control system for agile badminton robots. 
We harmonize model-based and learning-based methods for whole body control and arm policy training. 
For whole body control, we employ model-based strategies for chassis movement and learning-based strategies for arm control. 
This decoupling simplifies arm control training and facilitates zero-shot policy transfer from simulation to reality. 
In additional, arm policy does not need retraining when switching chassis, aiding future hardware adaptation. 
For arm policy training, we introduce a physics-informed ``IL+RL'' training recipe. 
We develop a model-based strategy with privileged information to guide arm policy training during both IL and RL phases. 
To mitigate performance drop issue during the transition from IL to RL, we additionally trained the critic during the IL phase. 
Our training scheme eliminates the need for complex reward shaping or curriculum learning, allowing training with sparse rewards.
And soft boundaries built by model-based policy supervision make policy network exploration in RL more efficient and safe.
Our main contributions are:
\begin{itemize}[itemsep=1pt,topsep=1pt,leftmargin=12pt]
\item We develop \ourmethod, the first whole body control system for agile badminton robots that integrates model-based and learning-based control strategies.
\item We propose the physics-informed training pipeline, including a model-based strategy with privileged information to warm up the actor and critic during imitation learning phase, and supervise further policy exploration during reinforcement learning phase.
\item We validate our framework in real-world environments and demonstrate that \ourmethod~achieves zero-shot generalization to multiple chassis without arm policy re-training.
\item Overall, \ourmethod~achieves 94.5$\%$ success rate against a badminton serving machine, while 90.7$\%$ success rate against human players, with maximum rally length of 40.
\end{itemize}

Our proposed system can be well generalized to other mobile manipulation tasks, especially those related to agile robot control, such as high speed object catching.

\section{Preliminary}

\subsection{Badminton Robot Architecture}
The badminton robot consists of a robotic arm and a high-speed chassis (see Fig.~\ref{whole system figure in real world}).
The robotic arm holding a badminton racket at its end has 5 rotating joints whose angular position and velocity can be controlled.
The high-speed chassis is able to move in all directions with a maximum speed of 5 $m/s$.
Visual perception relies on high frame rate binocular cameras behind the robot.

\subsection{Model-based Control Strategy for Badminton Robots}
The model-based strategy used to control the robot in this work is suggested by the work \cite{yang2022varsm}.
We denote the robot configuration vector by $\mathit{q \in \mathbb{R}^7}$, where $\mathit{q_{1:2} \in \mathbb{R}^2}$ represents the chassis's planar position in court, and $\mathit{q_{3:7} \in \mathbb{R}^5}$ represents the joint angles of the robotic arm.
Inverse kinematics function $\mathit{IK(p_r, R_r, v_r) = (q, \dot{q})}$ resolves joints' angle $\mathit{q}$ and velocity $\mathit{\dot{q}}$ from the racket's position $\mathit{p_r \in \mathbb{R}^3}$, orientation $\mathit{R_r \in SO(3)}$ and velocity $\mathit{v_r \in \mathbb{R}^3}$.
$\mathit{F_o}$ maps time $\mathit{t}$ to detected ball's position $\mathit{p_o(t)}$ and velocity $\mathit{v_o(t)}$.
$\mathit{\hat{F}_o}$ predicts ball's future position $\mathit{\hat{p}_o(t;\theta)}$ and velocity $\mathit{\hat{v}_o(t;\theta)}$ with a set of learnable parameters $\mathit{\theta}$.
For the process of hitting a ball, we first identify the trajectory of the shuttlecock, generating $\mathit{F_o}$ and $\mathit{\hat{F}_o}$. 
Using $\mathit{\hat{F}_o}$, we determine the appropriate arrival position ($\mathit{p_h}$) and hit time ($\mathit{t_h}$), where $\mathit{p_h=\hat{p}_o (t_h;\theta)}$. 
Subsequently, we use $\mathit{\hat{v}_o(t_h)}$ to calculate the racket orientation ($\mathit{R_h}$) and velocity ($\mathit{v_h}$) at $\mathit{t_h}$. 
Desired $\mathit{\hat{q}(t_h)}$ and $\mathit{\dot{\hat{q}}(t_h)}$ are acquired through $\mathit{IK}$.
Lastly, a low-level controller coordinates the joints to hit the ball.

Model-based methods offer stable hitting but need an accurate dynamic model and struggle in complex environments. 
In contrast, learning-based methods are more adaptable but hard to apply to complex tasks. 
Can we combine both to achieve whole-body control of the badminton robot?
\section{\ourmethod: \underline{Ha}rmonizing \underline{Le}arning-Based \\ Manipulation with \underline{M}odel-Based Locomo\underline{t}ion for \\ Whole-Body Badminton Robot Control}

\subsection{Whole-Body Control Strategy Architecture}
\begin{figure*}[t]
\centering
\includegraphics[width=1\linewidth]{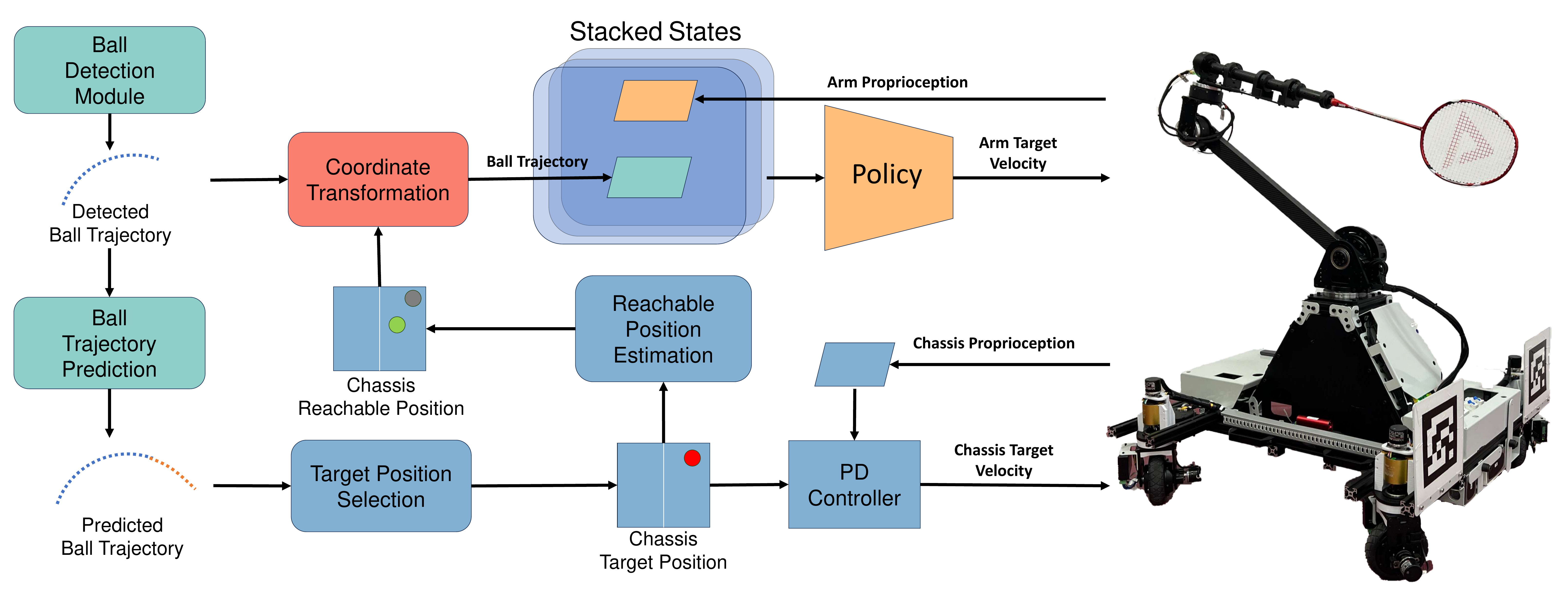}
\vspace{-8mm}
\caption{Overview of the components of \ourmethod.The green boxes represent the processing steps, including the detection and prediction of the ball trajectory. The blue boxes represent the model-based strategy for controlling the chassis, as discussed in \ref{chassis control policy}. The dark orange box represents the rigid transformation of ball trajectory coordinates, as discussed in \ref{itegrating method}. The yellow box represents the learning-based policy for controlling the robotic arm, as discussed in \ref{robotic arm control policy}.}

\label{method_system_figure}
\end{figure*}

The robot controlled by our system has two parts: the chassis and the arm. 
The chassis moves fast and is more prone to dangerous behavior, making it unsafe for black box network control. 
We also find that the chassis is harder to control than the robotic arm in sim2real (see Sec.~\ref{subsection:sim2realGap}), complicating the deployment of learning methods. For these reasons, we use a model-based method for the chassis and a learning-based method for the robotic arm (see Fig.~\ref{method_system_figure}).

\subsubsection{Overview} \label{itegrating method}
We define the ball hitting process so that the chassis provides a rough initial position $\mathit{p_{base}}$, and the arm performs fine actions at $\mathit{p_{base}}$. 
When moving to the target position $\mathit{p_{tar}}$, we use the low-level controller parameters and the current motion state of the chassis to estimate the reachable position as $\mathit{p_{base}}$, and sends it to the arm policy.
This means that the arm policy operates on the premise that the robot has reached $\mathit{p_{base}}$.
The arm policy does not directly use $\mathit{p_{base}}$ as input; rather, it integrates the ball trajectory data. 
The ball trajectory $\mathit{F}_o$ is translated from the world to the robot coordinate system originating at $\mathit{p_{base}}$, with an unchanged coordinate axis orientation.
Given the robot’s unavoidable deflection at high speeds, in addition to translating the trajectory using $\mathit{p_{base}}$, we rotate it corresponding to the robot’s orientation $\mathit{\alpha}$. 
The process results in a rigid transformation of the ball trajectory:
\begin{equation} \label{widetilde_p_o}
\mathit{ \widetilde{p}_o = R_o \cdot p_o^T + T_o },
\end{equation}
where $\mathit{\widetilde{p}_o}$ denotes the transformed ball trajectory.
$\mathit{T_o}$ is the translation vector calculated by $\mathit{p_{base}}$.
$\mathit{R_o}$ is the rotation matrix calculated by $\mathit{\alpha}$.

\subsubsection{Chassis Control Strategy} \label{chassis control policy}
We first determine the remaining action execution time as $\mathit{t_{avail} = t_h - t_{cur}}$, where $\mathit{t_{cur}}$ is the current time. 
We then segment the selection of $\mathit{p_{tar}}$ based on $\mathit{t_{avail}}$.
When $\mathit{t_{avail} > T_b}$, we use a fixed hitting height $\mathit{H_b}$ within the ball trajectory representation $\mathit{\hat{F}_o}$ to select $\mathit{p_h}$, and compute $\mathit{q_{1:2}}$ as $p_{tar}$ using $\mathit{IK}$. 
Due to $\mathit{\hat{F}_o}$'s inaccuracy when short ball trajectory are detected, calculating $\mathit{p_{tar}}$ with $\mathit{H_b}$ yields a more stable motion for the chassis.
In addition, we use a trust factor $\sigma$ to further make the chassis more stable in the initial stage:
\begin{equation}
    \sigma = \min(\frac{t_{cur}}{T_{init}}, 1),
\end{equation}
where $T_{init}$ is a constant indicating the time length of the initial stage.
Thus, the target position for the controlled movement of the chassis is:
\begin{equation}
    \hat{p}_{tar} = p_{init} + \sigma (p_{tar} - p_{init}),
\end{equation}
where $p_{init}$ represents the initial position of the chassis.
As $\mathit{\hat{F}_o}$ is updated, $\mathit{p_{tar}}$ is recalculated continuously.
When $\mathit{t_{avail} <= T_b}$, there's insufficient time for the chassis to adjust to significant changes in target position, so $\mathit{p_{tar}}$ is no longer recalculated to maintain the stability of the platform's motion.

\subsubsection{Arm Control Policy} \label{robotic arm control policy}

The badminton task can be formulated as a Markov Decision Process (MDP), which consists of a state space $\mathcal{S}$, action space $\mathcal{A}$, reward function $\mathcal{R: S \times A \to \mathbb{R}}$, and dynamics function $\mathcal{P: S \times A \to S}$. We parameterize a policy $\mathcal{\pi : S \to A}$ as a neural network with parameters $\theta \in \Theta$, denoted as $\pi_\theta$.

The policy network we use for the arm is a four-layer MLP.
The MLP receives a tensor of shape $(N, S+3)$ as input, where $N$ is the number of past observations, and $S$ corresponds to the joint angle measurements of the $S$ $DOF$ arm.
The ``+3'' corresponds to the 3D position of the ball $\widetilde{p}_o$ as mentioned in \ref{itegrating method}.
In our experiments, $S = 5$ and $N = 8$, so the input has shape (8, 8).
The output of the MLP is a tensor with length $S$, which represents the velocity control command for each joint of the arm.

\subsection{Physics-Informed “IL+RL” Policy Training Method} \label{arm policy training}
We construct a physics-informed training recipe consisting of ``IL+RL'' to train $\pi_\theta$.
Specifically, a model-based control strategy $\hat{\pi}_{PM}$ access to privileged information is developed based on \cite{yang2022varsm} in simulation at first.
$\hat{\pi}_{PM}$ serves as a teaching guide for $\pi_\theta$ and a critic model $V_\phi$ training via IL.
Subsequently, we advance the training of $\pi_\theta$ via RL supervised by $\hat{\pi}_{PM}$. An overview of the training recipe is give in Algorithm \ref{arm policy training pseudocodes} and details are discussed below.

\begin{algorithm} [tbp]
%\textsl{}\setstretch{1.8}
\renewcommand{\algorithmicrequire}{\textbf{Input:}}
\renewcommand{\algorithmicensure}{\textbf{Output:}}
\caption{Physics-informed ``IL+RL'' training recipe}
\label{arm policy training pseudocodes}
\begin{algorithmic}[1]
\REQUIRE $\hat{\pi}_{PM}, \lambda$, M, Q
\STATE Initialize policy $\pi_\theta$, critic $V_\phi$
\STATE $//$ Phase 1: Imitation Learning
\FOR{i= 1 \textbf{to} M}
    \STATE Collect $\mathit{(s_t, a_t, r_t, s_{t+1} ) \sim \pi_\theta}$.
    \STATE Update $\theta$ by a gradient method w.r.t $J^{SUP}(\theta)$. (\ref{J_supvise})
    \STATE Update $\phi$ by a gradient method w.r.t $L_{BL}(\phi)$. (\ref{V_training})
\ENDFOR
\STATE $//$ Phase 2: Reinforcement Learning
\FOR{i= 1 \textbf{to} Q}
    \STATE Collect $\mathit{(s_t, a_t, r_t, s_{t+1} ) \sim \pi_\theta}$.
    \STATE Estimate advantages $\hat{A}_t$. (\ref{estimate A})
    \STATE Update $\theta$ by a gradient method w.r.t $J^{RL}(\theta)$. (\ref{J_RL})
    \STATE Update $\phi$ by a gradient method w.r.t $L_{BL}(\phi)$. (\ref{V_training})
\ENDFOR
\ENSURE $\pi_\theta$
\end{algorithmic}  
\end{algorithm}

\subsubsection{Preparations}
To enhance the training of $\pi_\theta$, we have made the following preparations:
\begin{itemize}[itemsep=1pt,topsep=1pt,leftmargin=12pt]
\item \textbf{Teacher access to privileged Information  :} In simulation, we gain access to privileged information that is difficult to obtain in the real world and the completed badminton trajectory is one of them. 
We develop a model-based strategy $\hat{\pi}_{PM}$, has direct access to the complete badminton trajectory, to minimize control errors caused by ball trajectory prediction.
We leverage $\hat{\pi}_{PM}$ to supervise policy network training in both IL and RL phase.
\item \textbf{Augmented Real-world Ball Trajectories :} For training data, we map real-world badminton trajectories into the simulation to narrow the gap between simulation and reality. 
To address the sparsity in these trajectories, we apply translation and flipping to them.
This data augmentation technique not only expands the training dataset but also ensures an even distribution of ball landing points across the court's left and right halves.
\item \textbf{Chassis Control Strategy with Noise :} Throughout training, we consistently employ the same control strategy for the chassis as used in the real robot.
To minimize the gap between simulation and reality and achieve zero-shot transfer from simulation to reality in the policy network, we inject noise into the control of the simulated environment chassis.
\end{itemize}

\subsubsection{IL Phase} \label{IL phase}
In order to speed up the training of the policy, we use IL for a warm-up of $\pi_\theta$.
To avoid covariance drift problems, we use Dataset Aggregation (DAgger) \cite{ross2011dagger} as the algorithm of the IL phase.
In this phase, our optimization objective is to maximize :
\begin{equation} \label{J_supvise}
    \mathit{J^{SUP}(\theta) = \mathbb{E}_{s_t \sim \pi_\theta}[\ln{\pi_\theta(\hat{\pi}_{PM}(s_t) | s_t)}]}.
\end{equation}
In order to meet the needs of a actor model and a critic model for later RL training, we make some modifications to DAgger.
While we train $\pi_\theta$, we also use sampled data to train a critic model $V_\phi$ by minimize :
\begin{equation} \label{V_training}
    L_{BL}(\phi)=-\sum_{t=1}^T(\sum_{t^{'} > t}\gamma^{t^{'}-t} r_{t^{'}} - V_\phi)^2,
\end{equation}
where $\gamma$ is a discount factor. 
During the policy training, not only the data of correct actions will be generated, but also some data of wrong actions will be obtained.
And the data with balanced positive and negative samples is very beneficial for the critic model's training \cite{lu2021aw}.

\subsubsection{RL Phase} \label{RL phase}
In order to further improve the performance of $\pi_\theta$, we use RL to allow $\pi_\theta$ to further explore to achieve a higher success rate.
We use Proximal Policy Optimization (PPO) \cite{schulman2017ppo} as the RL algorithm with several modifications.
With pre-trained critic $V_\phi$ and the discount factor $\gamma$ , we estimate the advantages:
\begin{equation} \label{estimate A}
    \hat{A}_t = \sum_{t^{'} > t} \gamma^{t^{'}-t} r_{t^{'}} - V_\phi.
\end{equation}
The optimization objective of PPO is to maximize :
\begin{equation}
\begin{aligned}
    \mathit{J^{PPO}(\theta) =}& \\
    \mathit{\mathbb{E}_t [\min{(p_t(\theta) \hat{A}_t,}}
    \mathit{clip(p_t(\theta),} & \mathit{1-\epsilon, 1+\epsilon)\hat{A}_t)]},
\end{aligned}
\end{equation}
where $p_t(\theta)= \frac{\pi_\theta(a_t|s_t)}{\pi_{\theta_{old}}(a_t|s_t)}$.
In order to avoid excessive bias in the exploration of $\pi_\theta$, supervision from $\hat{\pi}_{PM}$ denoted as $J^{SUP}(\theta)$ is integrated into the PPO optimization objective with a weight of $\lambda$.
Therefore, the optimization objective in our RL phase is to maximize :
\begin{equation} \label{J_RL}
    \mathit{J^{RL}(\theta) = J^{PPO}(\theta) + \lambda J^{SUP}{(\theta)}},
\end{equation}
where $\lambda$ is a The reward function we used during the RL training is designed as follows:

\begin{itemize}[itemsep=1pt,topsep=1pt,leftmargin=12pt]
\item  \textbf{Hitting and Success Reward:} A sparse reward will be given when the racket touches / returns the ball;
\item \textbf{Penalties for Exceeding Dynamic Constraints:} A penalty is given when the robot's angle / velocity / acceleration / jerk surpass predefined limits in an episode.
\item \textbf{Penalties for Unsafe Actions:} A penalty is given when the base joint angle of the robotic arm is too backward / the robot self collides / the robot collides with other objects / the racket is below a certain height.
\item \textbf{Ball Position Reward:} A dense reward based on the ball's closest distance to the paddle's center, with smaller distances yielding higher rewards.
\item \textbf{Net Crossing Reward:} A sparse reward is given when the robot hits the ball over the net.
\item \textbf{Ball Height above Net Reward:} A dense reward based on the ball's height when crossing the net.
\item \textbf{Landing Position Reward:} A dense reward calculated by the distance from the center of the opponent's court to where the ball lands after being hit back.
\end{itemize}

\textbf{Remark}: 
The superiority of our proposed learning-based policy over the model-based strategy for robotic arm control can be attributed to the following reasons:

\begin{itemize}[itemsep=1pt,topsep=1pt,leftmargin=12pt]
\item[i)] \textbf{Higher Upper Bound in IL Phase}: The model-based strategy $\hat{\pi}_{PM}$ access to privileged information provides a higher performance upper bound than the real model-based strategy $\pi_{PM}$ during the IL phase.
\item[ii)] \textbf{Greater Potential in RL Phase}: The explorative training in the learning-based policy during the RL phase provides an opportunity to learn actions beyond the capabilities of model-based strategies.
\end{itemize}
\section{Experiments}
\begin{figure}[t]
\centering
\subfloat[]
{
    \includegraphics[height=3.5cm]{figures/camera.pdf}
}
\hspace{1cm}
\subfloat[]
{   
    \includegraphics[height=3.5cm]{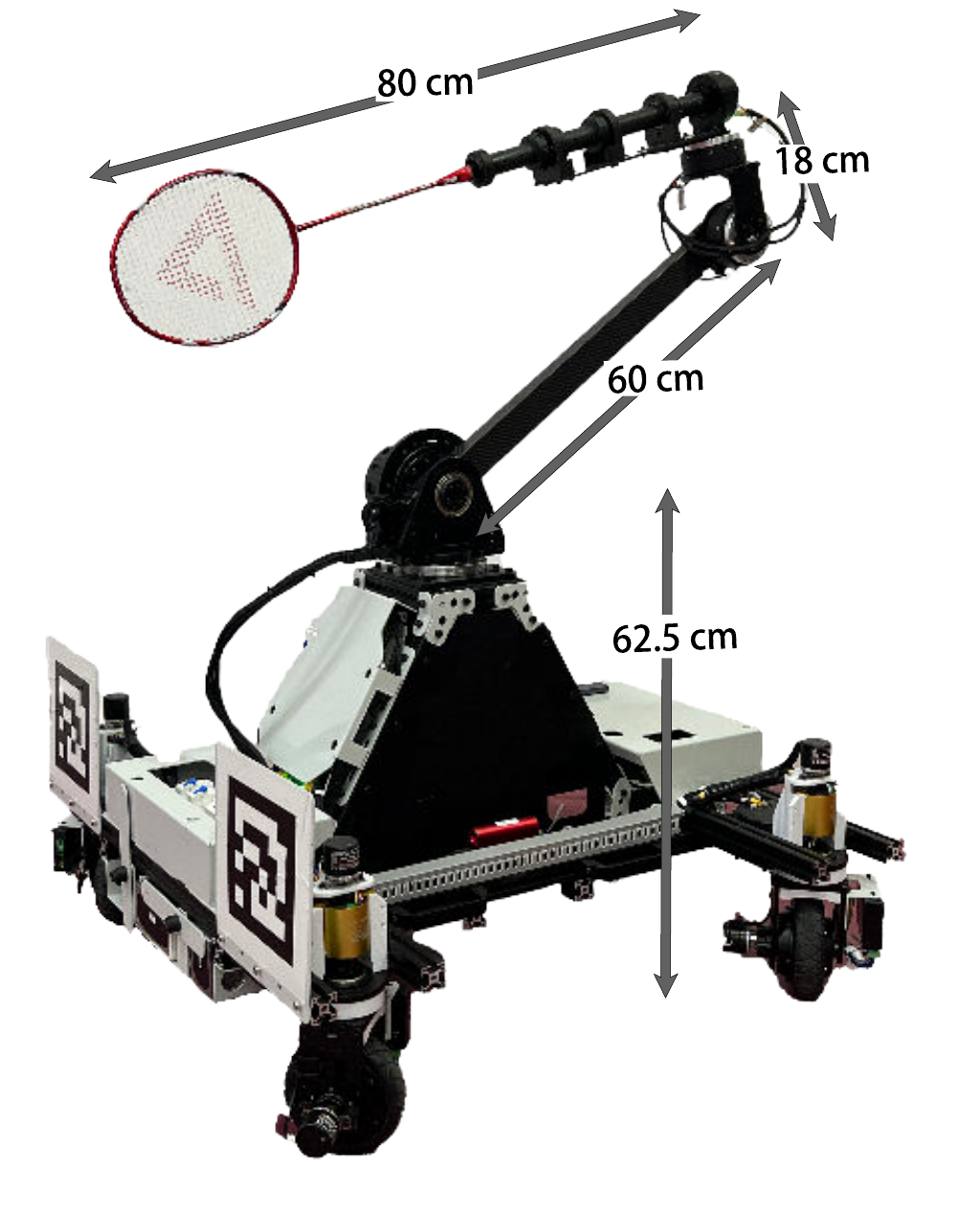}
}
\caption{\textbf{Hardware of badminton robot system.} Our badminton robot system consists of two parts: (a) vision module and (b) robot body.The visual module handles scene perception, including ball recognition, tracking, and robot positioning. The computer on the robot body executes the control algorithm and send control commands to each motor.}
\label{hardware fig}
\end {figure}

We evaluate our system in both simulation and real-world.
The simulation environment is built using PyBullet \cite{coumans2016pybullet}.
In the real world, the hardware of the robot we used is shown in Fig.~\ref{hardware fig}. 
It has two parts: a binocular vision module and a robot body.
We gather 10,000 raw badminton trajectories in the real-world.
And we use 9,000 trajectories enhanced through data augmentation to train the learning-based policy network in simulation, while the remaining 1,000 served as the test dataset.
Experiments are conducted in the real world with our self-engineered badminton robot against both a badminton serving machine and human.

The metrics we use to evaluate our system is as follows:
\begin{itemize}[itemsep=1pt,topsep=1pt,leftmargin=12pt]
\item \textbf{Hit Rate}: the rate of the robot racket touching the ball.
\item \textbf{Success Rate}: the rate of the robot successfully hitting the ball to the designated area on the opponent's court.
\item \textbf{Max Rally}: the maximum number of consecutive rally rounds between the robot and the human.
\end{itemize}

\begin{figure}[t]
\centering
\includegraphics[width=0.8\linewidth]{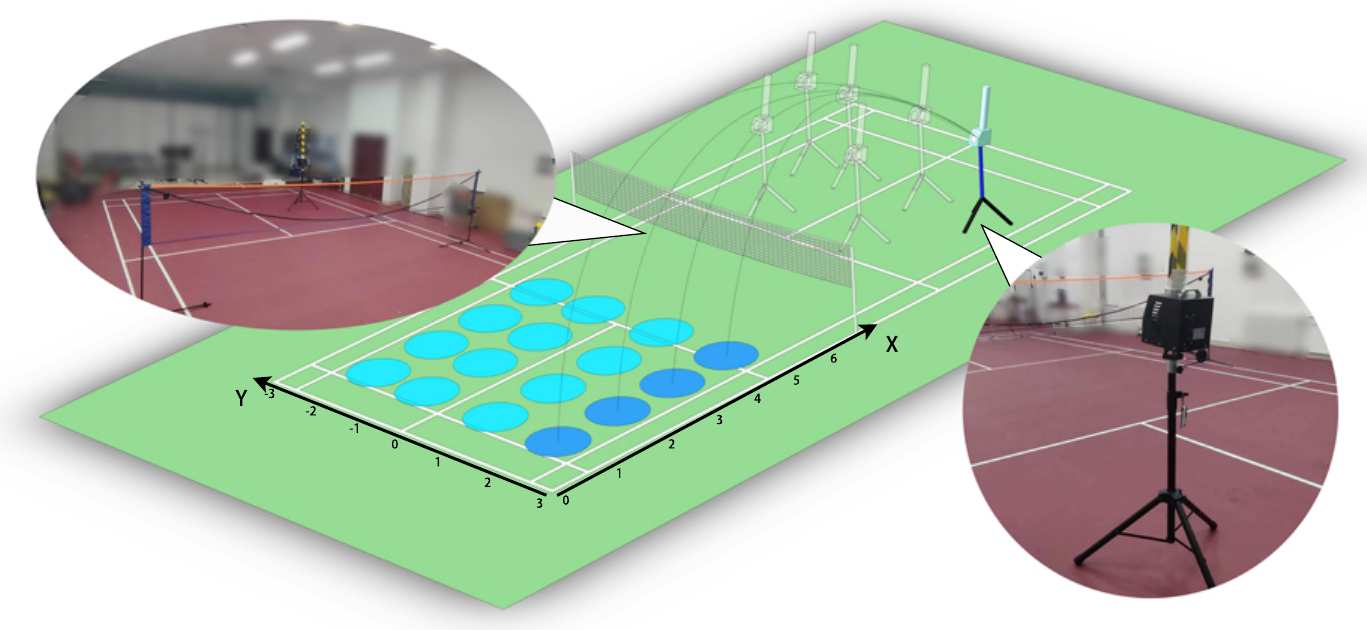}
\caption{\textbf{Serving machine experiments setup.} 
We set 6 positions for the serving machine on the right side of the court. We configure 20 combinations of machine positions, power, and angle to ensure diverse trajectories. The circles on the left side of the court show the ball landing points. The subplots on the left and right show different views of the real court.
}
\label{serving_machine_scenarios}
\end{figure}

\subsection{Gap Analysis from Simulation to Reality} \label{subsection:sim2realGap}
\begin{figure}[t]
\centering
\subfloat[]{
\begin{minipage}[b]{.45\linewidth}
    \centering
    \includegraphics[width=\linewidth]{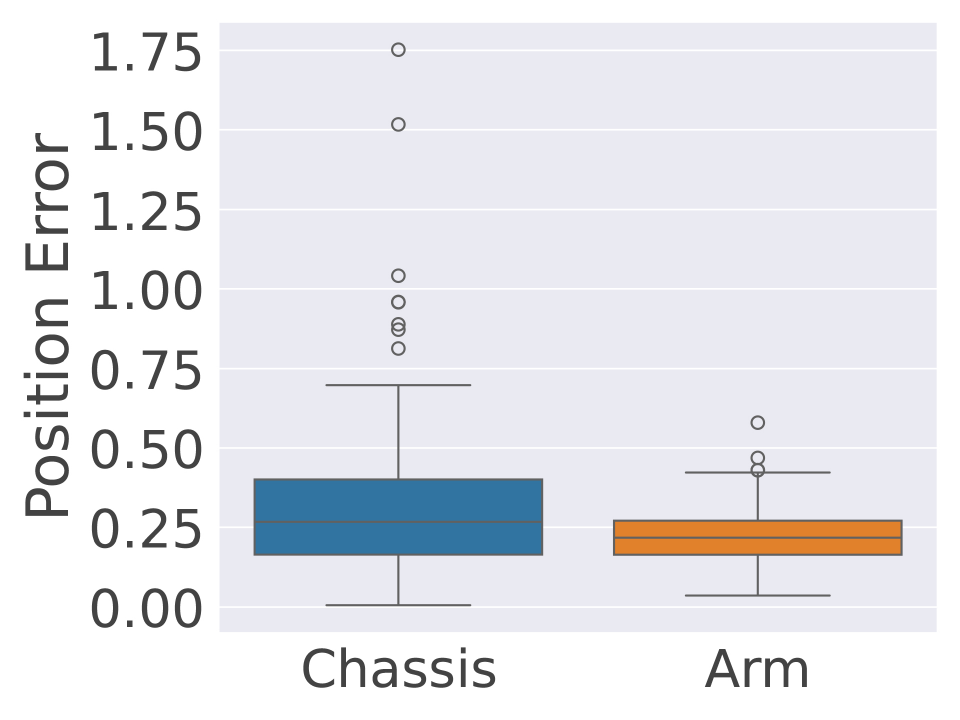}
    \end{minipage}
}
\hfill
\subfloat[]{
\begin{minipage}[b]{.45\linewidth}
    \centering
    \includegraphics[width=\linewidth]{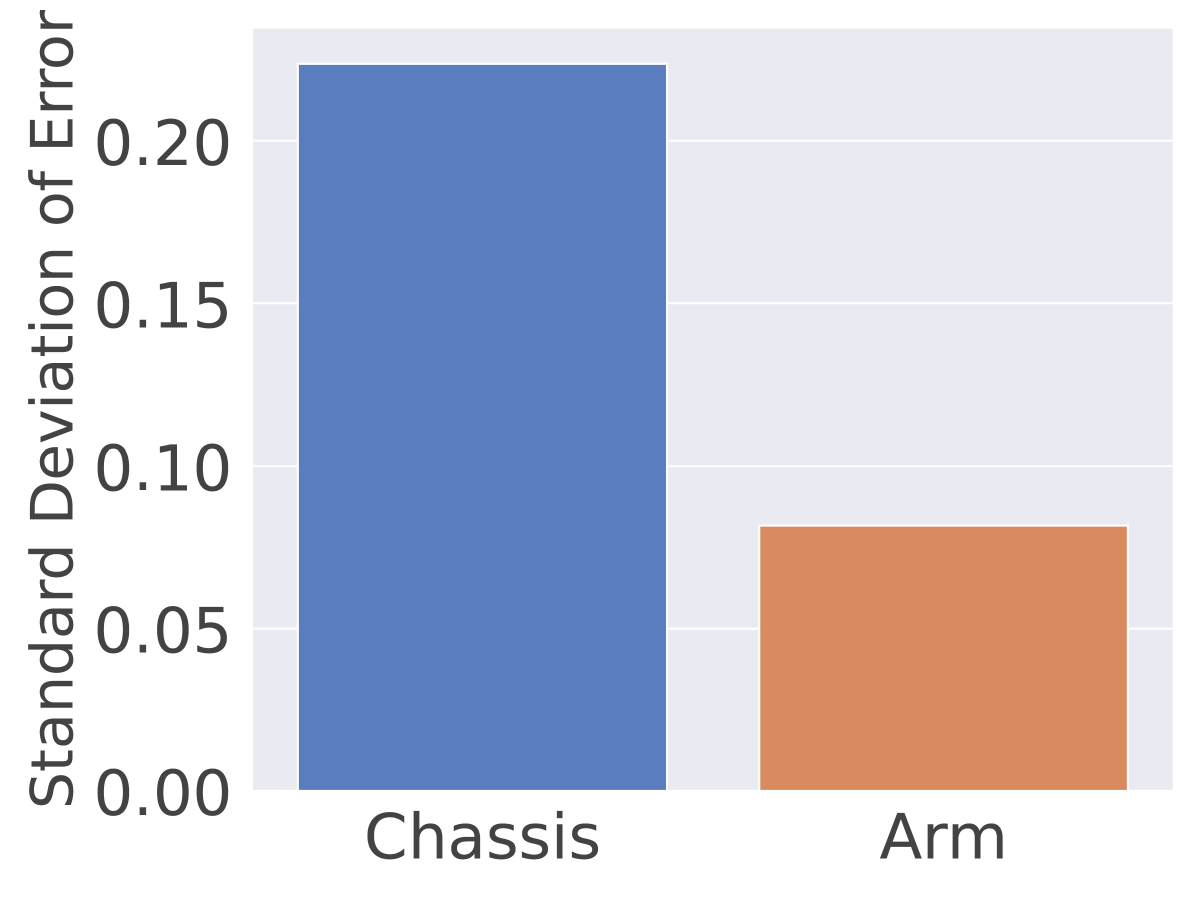}
    \end{minipage}
}
\vspace{-2mm}
\caption{\textbf{Comparison of the control gap between simulation and real world of chassis and arm.} We test the difference between the mobile platform and the robot's end position in the simulation and real world under the same control commands. Box plot (a) visualizes the end distance error. We show the standard deviation of the end distance error for the chassis and arm in chart (b). The results show that the chassis is more volatile to changes in the environment and is less conducive to sim2real.}
\label{fig:sim2realGap}
\end{figure}

We test the difference between simulation and reality for the robot's chassis and manipulator separately, using the same 194 control trajectories in both environments. 
The difference for the chassis is measured by its final position, while for the manipulator, it is the final position of the racket. 
Results in Fig.~\ref{fig:sim2realGap} show the manipulator's gap is smaller, with the chassis error showing more variation. 
This may be because the chassis must contact the ground, involving more complex physical processes and varying contact conditions. In contrast, controlling the robot arm is relatively simpler.
This suggests that chassis control is less stable and more affected by real-world conditions, which is why we use a learning method for the manipulator and a model-based method for the chassis.

\subsection{Model-based Strategy vs. Learning-based Policy}
We compare the performance of the model-based strategy and our learning-based policy for the arm control in real world. 
The model-based strategy achieve a hit rate of 91\% and a success rate of 77.5\%, while our learning-based policy attained a hit rate of 97\% and a success rate of 94.5\%. 
It reveals that the learning-based policy outperforms the model-based strategy in both hit and success rates, with a notably higher success rate.

\subsection{Ablation Study}
\begin{figure}[t]
\centering
\includegraphics[width=0.7\linewidth]{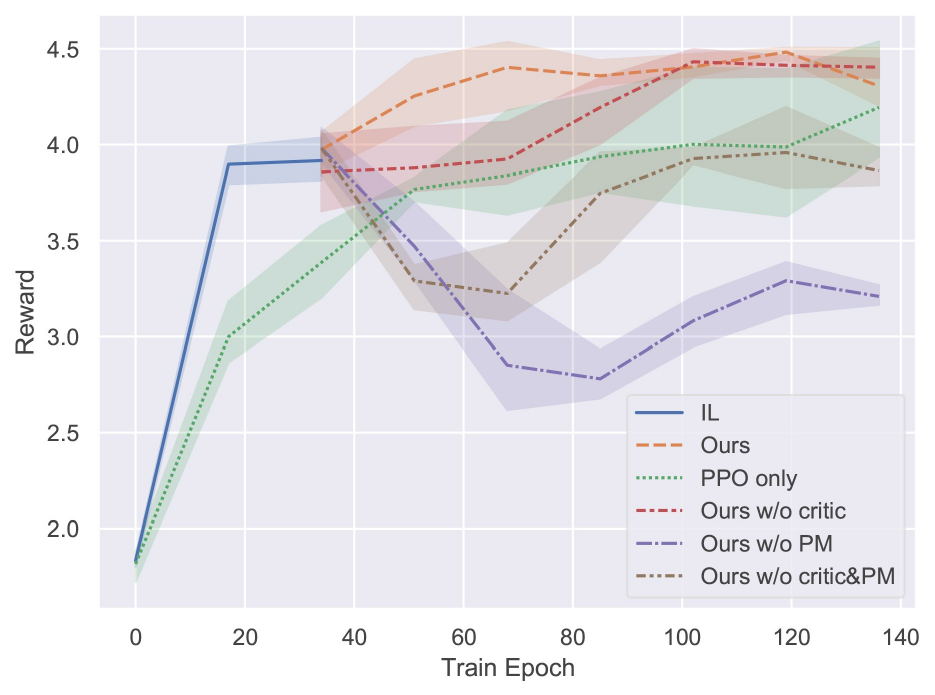}
\vspace{-3mm}
\caption{\textbf{Ablation Study.} We sequentially strip components from our proposed training recipe and observe the ensuing reward fluctuations during training. \textbf{IL}: The IL phase of our training recipe detailed in \ref{IL phase}. \textbf{Ours}: The RL phase of our training recipe detailed in \ref{RL phase}. \textbf{PPO only}: Employ solely PPO to train the policy network and learn from scratch. \textbf{Ours w/o critic}: The omission of critic warm-up. \textbf{Ours w/o PM}: The absence of model-based strategy supervision during RL. \textbf{Ours w/o critic\&PM}: Neither critic warm-up nor model-based strategy supervision in the RL phase.}
\label{il_rl_training}
\end{figure}

We conduct ablation study on our proposed arm policy training recipe. 
As shown in Fig.~\ref{il_rl_training}, the inclusion of the IL phase can rapidly enhance the performance of the model compared to initiating RL from scratch. 
Our training recipe demonstrates further optimization of the policy network's performance during the RL phase.
Without the supervision of a pre-trained critic or model-based strategy, policy networks experience a significant initial drop in reward during reinforcement learning, undoubtedly extending the exploration phase. 
Notably, the lack of supervision from a model-based strategy exacerbates this drop in reward and leads to greater fluctuations.
Notably, we find that the absence of model-based policy supervision exacerbates reward degradation and leads to greater volatility when the policy network is trained.

\subsection{Robot vs. Serving Machine} \label{against serving machine}
We configure the serving machine using combinations of three positions, three angles, and two strengths, creating 20 scenarios (see Fig.~\ref{serving_machine_scenarios}). 

For each scenario, we collect 10 test outcomes for each of the model-based and learning-based strategies, resulting in 200 outcomes per strategy.

As shown in Tab.~\ref{tab:against servering machine}, the hit rate and success rate of robots deployed with the learning-based policy is significantly better than the model-based strategy.

\begin{table}[t]
\caption{Comparison Results Between Model-Based Policy and Ours in Real World}
\vspace{-3mm}
\centering
{
\begin{tabular}{@{}ccc@{}}
\toprule
              & Hit Rate & Success Rate \\ \midrule
Model-Based   & 91.0       & 77.5 \\
Ours   & \textbf{97.0} & \textbf{94.5}\\ \bottomrule
\end{tabular}
}
\label{tab:against servering machine}
\end{table}

\subsection{Human Robot Interaction} \label{against human}
\begin{figure}[t]
\centering
\includegraphics[width=0.8\linewidth]{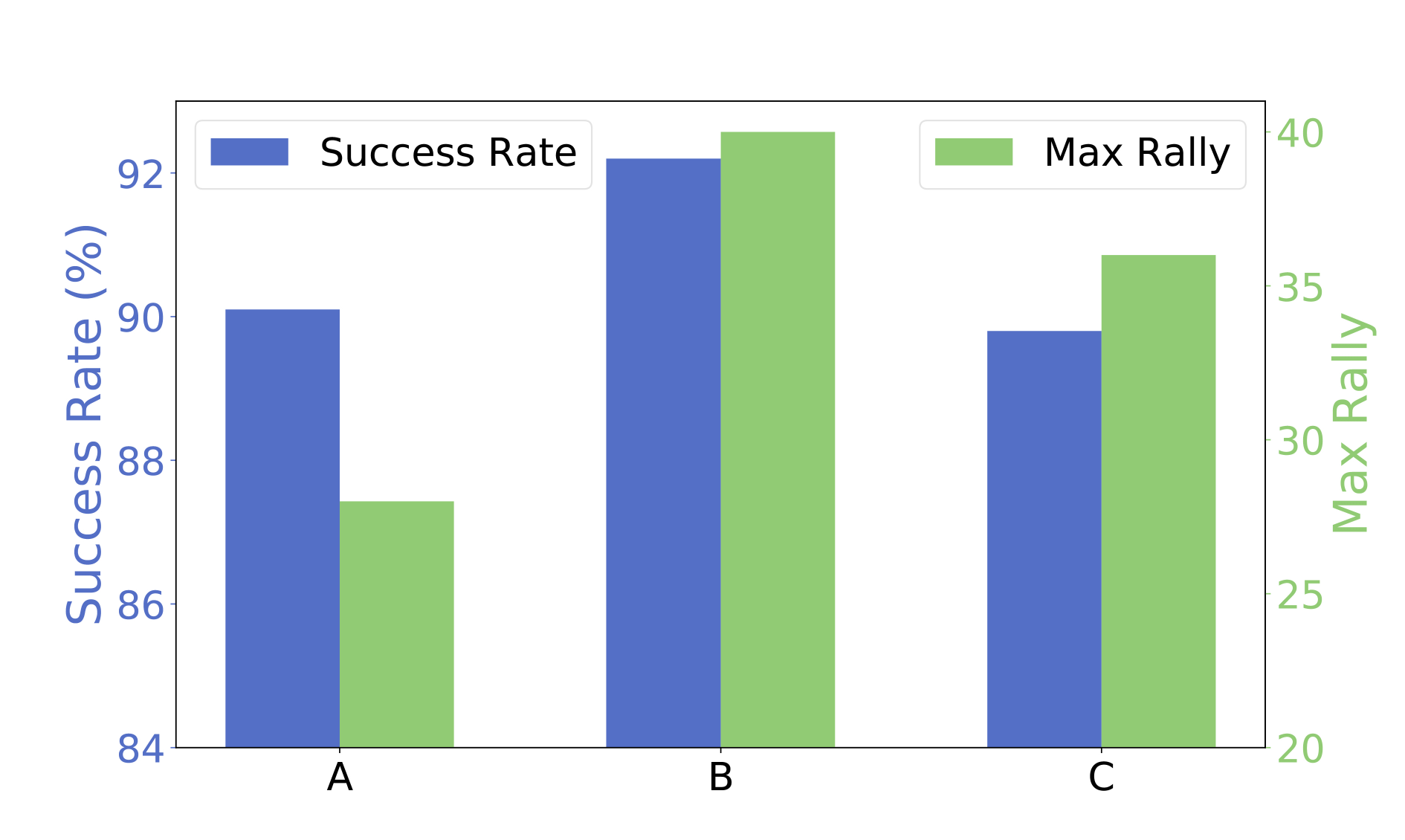}
\vspace{-5mm}
\caption{\textbf{Robot vs. Human Performance.} We collect three sets of data from robot-human matches. In each set, human and robot play 100 rounds, and we record the robot's return success rate and the maximum rally. We find that our system performs well during play with humans.}
\label{against_human_result}
\end{figure}

To verify the performance of the robot and human in continuous combat, we collected three sets of human-robot combat data. 
Each set of data contains 100 rounds of battles between humans and robots.
We recorded the robot's return success rate and the longest round in the game with humans, as shown in Fig.~\ref{against_human_result}.
We found that our method still showed good return results in the process of playing against humans, with an average return success rate of 90.7$\%$, and could play up to 40 rounds with human players.

\section{Related Work}
\subsection{Sports Robots}
As far as we know, there have been some studies on ball sports robots.
Xiao et al.~\cite{xiao2024gatch} and Voeikov et al.~\cite{voeikov2020ttnet} proposed an information perception method for ball sports scenes.
Google team proposed a table tennis robot \cite{abeyruwan2023sim2real,ding2022learning,d2023robotic,gao2020robotic,d2024achieving}.
They used model-free reinforcement learning algorithms to complete the learning of the robot's control policy network. 
Ji et al.~\cite{ji2021model} used a self-designed six-degree-of-freedom robot system to realize the table tennis return task through a control method based on a physical model.
Xiong et al.~\cite{xiong2012impedance} implemented two humanoid robots to complete the table tennis pulling task.
Zaidi et al.~\cite{zaidi2023athletic} proposed the first open source autonomous wheelchair robot for regular tennis matches, establishing the first baseline for replicable wheelchair robots used in regular singles matches.
In our system, The robot that is controlled has a more flexible joint setup.
And to the best of our knowledge, our system is the first to integrate model-based and learning-based control of an agile badminton robot.

\subsection{Integration of RL and IL}
There are some works that combine IL and RL to speed up training while reaching ultra-high levels beyond humans. 
Ramrakhya et al.~\cite{ramrakhya2023pirlnav} proposed a two-stage learning scheme of ``IL+RL'' on navigation tasks.
Nair et al.~\cite{nair2020awac} provides an integrated framework combining IL and RL that can pre-train policies using large amounts of offline data and then quickly fine-tune them through online interactions to learn complex behaviors. 
Lu et al.~\cite{lu2021aw} proposed a training plan that balances positive and negative samples in critic model training and only uses successful data to update Actors, solving the problem of sudden performance drop in the initial stage of migration from imitation learning to reinforcement learning. 
Rajeswaran et al.~\cite{rajeswaran2017learning} used real-person operation data obtained from VR devices to enhance reinforcement learning policies to learn complex dexterous hand operation tasks.
In our ``IL+RL'' training recipe, a model-based strategy accessing privileged information guides and sets soft boundaries for policy network training, rather than producing offline data. 
We also warm up the critic model during IL phase, alleviating performance drop-off from IL to RL.

\subsection{Combined Model-Based and Learning-Based Control}

In recent years, more and more works analyze and combine the advantages of both and design integrated control algorithms.
Abeyruwan et al.~\cite{abeyruwan2023agile} compares the advantages and disadvantages of physical model-based policies and learning-based policies on the task of catching a high-speed flying ball. 
Dong et al.~\cite{dong2020catch} proposes a hierarchical control algorithm to complete the capturing task.
They use SQP and QP for motion trajectory planning and then use a learning-based controller to track the trajectory.
Pandala et al.~\cite{pandala2022robust} 
proposed a nonlinear controller based on virtual constraints and low-level quadratic programming, and used RL to train a neural network that computes the RMPC uncertainty set for a robust walking task on rough roads for Legged robots.
In our system, model-based and learning-based strategies are separately applied to varied components of the robotic control.
For the arm policy, the model-based strategy plays a guiding role, while the neural network provides accelerated running speed and facilitates easier deployment.
\section{Conclusion}
We present \ourmethod, a new whole-body control system for an agile badminton robot, combining model-based and learning-based control methods. 
Our system leverages model-based control for the chassis and learning-based control for the robot arm. 
The arm policy training employs an ``IL + RL'' strategy, where model-based techniques pre-train the actor and critic in IL, enhancing subsequent RL policy training. 
Our policy enables zero-shot transfer from simulation to reality and can seamlessly integrate into different chassis without requiring retraining. 
Extensive experiments demonstrate a 94.5{\%} success rate against a serving machine, 90.7{\%} against a human opponent, and up to 40 consecutive hits. Our system is robust, adaptable, and applicable to other agile robot control tasks.

\bibliographystyle{IEEEtran}
\bibliography{references}
\end{document}